\documentclass[journal]{IEEEtran}

\usepackage[utf8]{inputenc}
\usepackage[T1]{fontenc}
\usepackage{microtype}
\usepackage{booktabs}
\usepackage{array}
\usepackage{xcolor}
\usepackage{amsmath}
\usepackage{tikz}
\usepackage{pgfplots}
\pgfplotsset{compat=1.17}
\definecolor{okblue}{RGB}{0,114,178}
\definecolor{okorange}{RGB}{230,159,0}
\definecolor{okverm}{RGB}{213,94,0}
\definecolor{okgreen}{RGB}{0,158,115}
\usepackage[hidelinks]{hyperref}
\usepackage{xurl}
\usepackage{needspace}

\newcommand{\code}[1]{\texttt{#1}}
\title{The Integer Alibi: Localizing Cross-Kernel\\
Divergence in INT8-Quantized LLM Inference}

\author{Teng-Ruei~Chen%
\thanks{T.-R. Chen is with Krixvon, Taipei 100, Taiwan (e-mail: luka@krixvon.com; ORCID: 0000-0001-8995-334X).}%
\thanks{Preprint, August 2026; \texttt{arXiv:2608.13756}. All measurements ran on a pinned software stack (vLLM 0.27.1 by container digest, SGLang by container digest) on a single RTX 4090; the protocol was pre-registered before any backend-comparison measurement and every amendment is append-only. Artifacts, pre-registration, and per-experiment manifests are listed in the artifact statement.}}

\begin{document}
\maketitle

\begin{abstract}
Two GPU kernels implementing the same scaled INT8 GEMM interface are usually treated as interchangeable. We test that assumption: holding the checkpoint, prompts, hardware, inference engine, decoding, and quantization configuration fixed, we swap only the INT8 linear kernel (CUTLASS versus Triton) inside vLLM. At 1.7B each arm reproduces itself bit-for-bit across cold restarts, yet the arms agree on no sequence in any end-to-end comparison we ran (0/8, 0/16, and 0/64). What makes this more than a benchmark discrepancy is an \emph{integer alibi}: for shared INT8 operands under a verified no-overflow bound, the INT32 dot product is exact and order-independent, so the accumulator cannot be the source of any difference. Feeding both kernels identical operands from every linear layer of Qwen3-1.7B and 8B (196 and 252 layers), we find bit-identical outputs under power-of-two scales, confirming a pinned prediction list 196/196 and 252/252 (pre-registered at 1.7B, pinned but not blind at 8B), and observed differences of at most one bfloat16 spacing under the checkpoints' real scales. This localizes the divergence to scale application and output rounding after the exact accumulator. Applied as a probe checkpoint, the same intervention restores end-to-end bitwise agreement (8/8 and 16/16 sequences). Cross-implementation FP8 GEMM shows a different signature: both the prevalence and the magnitude of differences grow with reduction depth, while the INT8 fraction stays at parts per million and within one spacing over a 64$\times$ range of $K$. Teacher-forced replay ties layers to tokens: flips concentrate at small logit margins, which predict flip risk with ROC-AUC 0.94 on 16{,}384 positions. We will release the pre-registration, per-layer predictions, manifests with kernel-selection evidence, and a conformance procedure that turns these controls into a concrete check for kernel interchangeability.
\end{abstract}


\section{Introduction}
\label{sec:intro}

A serving engine that supports quantized inference typically carries several implementations of the same linear-layer interface: a vendor-library path, a compiler-generated path, and one or more fallbacks. Which one runs is decided at load time by hardware checks, environment variables, and heuristics; the choice is logged at debug level and is invisible in the model card, the benchmark table, and usually the paper. The implicit contract is that these implementations are interchangeable: same operands in, same answer out, up to negligible noise.

This paper tests that contract under unusually tight controls and finds that it fails in a specific, localizable way. On one GPU, one engine build, one INT8-quantized checkpoint, and one set of prompts, swapping a single kernel (vLLM's CUTLASS INT8 scaled-MM for its Triton counterpart, selected by one environment variable) changes every greedy continuation we compared, including all 64 in the largest run. Neither arm is noisy: at 1.7B each reproduced itself bit-for-bit across two cold restarts (we did not repeat the 8B or the 64-prompt runs). The two arms are simply two different deterministic functions.

The numerical behavior of tensor cores has itself been characterized directly~\cite{fasi2021tensorcores}, the FP8-versus-INT8 trade-off for inference has been studied at the format level~\cite{vanbaalen2023fp8int8}, integer-only inference and its calibration are long established~\cite{jacob2018integer,wu2020integer}, and floating-point non-associativity is a known source of irreproducibility in HPC and deep learning~\cite{shanmugavelu2024nonassociativity}, which floating-point work must address with reproducible-summation algorithms or fixed reduction trees~\cite{collange2015reproducible,ahrens2020reproducible} rather than by appeal to exactness, and the format landscape has broadened to microscaling blocks~\cite{rouhani2023mx}. Prior work has established the broader phenomenon that nominally deterministic inference varies across platforms~\cite{schlogl2023deviations}, batch sizes and GPU configurations~\cite{yuan2025nondeterminism,he2025batchinvariance}, tensor-parallel sizes~\cite{tbik2025}, and whole inference backends~\cite{pape2026silent}, and that discrepancies concentrate at kernel-boundary precision changes~\cite{zhu2026heal}. Our contribution is not the phenomenon but the isolation and the control. A backend change bundles kernels, caches, graph execution, and scheduler defaults; a hardware change bundles even more. Here the treatment is one kernel substitution inside a fixed engine, and the analysis rests on a control unavailable to floating-point studies:

\begin{quote}
For shared INT8 operands, if no dot product overflows INT32, the accumulator value is \emph{exact} and \emph{independent of reduction order}. Any cross-kernel difference must therefore originate after the accumulator, in scale application and output rounding.
\end{quote}

We call this the \emph{integer alibi}: under the verified bound, the accumulation stage has proof that it could not have committed the divergence. The alibi converts ``the kernels disagree'' into a falsifiable localization with a built-in negative control, and it enables an intervention: if the remaining suspects are the two floating-point scale multiplications, then making those multiplications exact (by rounding scales to powers of two, which commute with rounding under conditions we state and test) should restore bitwise equality. It does, at every layer and end to end.

Concretely, this paper contributes a bundle of controls and transfer measurements that, among the works we reviewed through August 13, 2026, we did not find previously combined; Section~\ref{sec:related} discusses the nearest concurrent results individually:

\begin{enumerate}
\item \textbf{Kernel-only treatment, audited where it can be.} Comparisons in which checkpoint, prompts, engine, decoding, and quantization configuration are fixed and the only declared difference is the linear kernel. For the final teacher-forced comparison the actually-selected kernel class is captured from execution logs into per-run manifests and verified by an identity contract; the earlier end-to-end token runs were the same intended one-variable intervention but carry no manifest-level kernel-selection evidence (Section~\ref{sec:methods}).
\item \textbf{The integer alibi as an exact control.} Per-layer predictions of where bitwise agreement is \emph{required}, derived from an exact fp64 emulation of the INT32 accumulator against the $2^{24}$ float32 representability threshold, confirmed 196/196 and 252/252. The 1.7B list was pre-registered; the 8B list was hash-pinned before the authoritative rerun but after an earlier result on the same data was known, making it a pinned rather than an independent prospective replication (Section~\ref{sec:layers}).
\item \textbf{Controlled epilogue localization.} With identical quantized operands, real-scale cross-kernel differences never exceeded one bfloat16 spacing at any of the 448 layers evaluated, and vanish under the power-of-two probe; boundary unit tests document exactly where the commutation argument fails (subnormals, overflow) (Sections~\ref{sec:layers}, \ref{sec:boundary}).
\item \textbf{A divergence-signature contrast.} Exact-accumulator INT8 versus float-accumulator FP8 across a 64$\times$ range of reduction depth: parts-per-million with no detectable trend in a single unreplicated sweep, versus percent-level and growing in both prevalence and magnitude, the latter compatible with square-root-shaped saturation under a model comparison with held-out folds (Section~\ref{sec:ksweep}).
\item \textbf{Layer-to-token transfer.} Teacher-forced replay separates single-step flips from autoregressive cascade; the logit margin predicts flips with ROC-AUC 0.942 (cluster CI over 64 prompts), giving a quantitative bridge from ulp-level differences to sequence-level disagreement (Section~\ref{sec:tokens}).
\end{enumerate}

We do not claim that deterministic inference is newly shown to be fragile, that INT8 makes whole-model inference bitwise reproducible, or that power-of-two scales are a validated deployment mitigation. Section~\ref{sec:limits} states these boundaries explicitly.

The conformance procedure we end with sits in an older tradition of differential and precision testing for numerical code~\cite{pham2019cradle,zhang2021predoo} and of equivalence checking for GPU kernels~\cite{dubey2025volta}, with the matrix-multiply semantics of successive tensor-core generations now formalized for SMT reasoning~\cite{valpey2025smt}; what we add is a set of checks specialized to the scaled-INT8 structure, five of the seven having falsifiable outcomes rather than thresholds to tune. The remainder of this paper is organized as follows. Section~\ref{sec:related} places the study among prior work by treatment variable. Section~\ref{sec:theory} derives the exactness structure that makes the integer alibi available, and states the conditions under which it fails. Section~\ref{sec:methods} describes the stack, the pre-registration, and the identity contract that makes ``only the kernel differs'' an audited property. Sections~\ref{sec:layers} through~\ref{sec:tokens} report the layer-level verification, the hardware boundary tests, the divergence signature versus reduction depth, and the transfer from layer differences to tokens. Section~\ref{sec:intervention} reports the power-of-two intervention end to end, Section~\ref{sec:discussion} extracts the conformance procedure, and Section~\ref{sec:limits} bounds what may be concluded.

\section{Related Work}
\label{sec:related}

Work relevant to this paper divides along one axis: \emph{what was actually varied} while everything else was held fixed. We organize it from the coarsest treatment variable (the platform) to the finest (a single kernel), noting within each group how the numerical path, the observation scale, and the available control differ from ours. Table~\ref{tab:related} summarizes the comparison; the text then follows each group chronologically.

\begin{table*}[t]
\caption{Prior work organized by treatment variable. The rightmost column states what this paper adds rather than what the prior work lacks; several of these studies isolate mechanisms we rely on.}
\label{tab:related}
\centering
\footnotesize
\begin{tabular}{@{}>{\raggedright\arraybackslash}p{2.5cm}>{\raggedright\arraybackslash}p{3.0cm}>{\raggedright\arraybackslash}p{2.2cm}>{\raggedright\arraybackslash}p{2.4cm}>{\raggedright\arraybackslash}p{4.4cm}@{}}
\toprule
Work & Treatment variable & Numeric path & Observation scale & Relation to this paper \\
\midrule
Schl\"ogl et al.~\cite{schlogl2023deviations} & Platform, SIMD, convolution algorithm & General NN inference & Layer, model output & Stacks are not numerically neutral; we fix the platform and isolate one kernel \\
Yuan et al.~\cite{yuan2025nondeterminism} & Batch size, GPU count/type, precision & BF16, FP32 & Token, accuracy, length & Greedy LLM outputs are configuration-sensitive; we fix configuration and add an exact integer control \\
He~\cite{he2025batchinvariance} & Batch size (serving shape) & Floating-point kernels & Kernel to token & Buys batch invariance with throughput; we study fixed-shape interchangeability \\
TBIK~\cite{tbik2025} & Tensor-parallel size & Floating-point reductions & Bitwise model output & Aligns reduction trees; the integer path needs only an epilogue contract \\
Pape et al.~\cite{pape2026silent} & Whole inference backend & Mostly floating-point serving & Benchmark score & Backend is a hidden hyperparameter; we isolate one kernel within it \\
HEAL~\cite{zhu2026heal} & Heterogeneous GPUs, 16-bit path & FP16/BF16 boundaries & Activation, token, task & Attributes error to kernel boundaries, uses teacher forcing; we add an exact-accumulator control and a scale intervention \\
Hawkeye~\cite{badash2026hawkeye} & GPU architecture, tensor-core semantics & FP16/BF16/FP8 & Instruction, matmul & Low-level arithmetic is auditable; we audit a serving-path GEMM and its transfer to tokens \\
Kernel contracts~\cite{xu2026contracts} & Training vs.\ inference kernels & Framework-level & Theoretical bounds & Supplies contract vocabulary; we give an executable instance \\
DiFR~\cite{karvonen2025difr} & Provider or implementation & Black-box or fingerprint & Token, activation & Detects deviation; we localize it between two named kernels \\
LiquidGEMM~\cite{liquidgemm2025}, Han and Qu~\cite{han2026matched} & Kernel, runtime, quantization pipeline & W4A8, INT4, FP16 & Throughput, latency & Ask which implementation is faster; we ask whether two agree numerically under fixed operands \\
\bottomrule
\end{tabular}
\end{table*}

\subsection{Platform and configuration}
Fixed weights and inputs do not imply numerically identical inference across deployment stacks. Schl\"ogl et al.~\cite{schlogl2023deviations} showed that CPU instruction selection and GPU convolution algorithm choices induce unanticipated output deviations across, and sometimes within, platforms. Yuan et al.~\cite{yuan2025nondeterminism} demonstrated that evaluation batch size, GPU count and type, and precision materially change greedy LLM outcomes, attributed the effect to floating-point non-associativity under varying reduction orders, and proposed computing in FP32 while storing in BF16.

\subsection{Execution regime and engine}
He~\cite{he2025batchinvariance} distinguished run-to-run determinism at a fixed shape from invariance across batch sizes, and built batch-invariant kernels at a measured throughput cost; TBIK~\cite{tbik2025} aligns reduction trees across tensor-parallel sizes for the same purpose. The Silent Hyperparameter~\cite{pape2026silent} holds weights, decoding, and hardware fixed and shows that swapping the whole inference backend shifts benchmark scores by up to double digits. Because an engine change bundles kernels, caches, graph execution, and defaults, it cannot isolate a mechanism, which is the gap our single-kernel treatment addresses.

\subsection{Kernel boundaries, audits, and contracts}
HEAL~\cite{zhu2026heal} attributes heterogeneous-GPU 16-bit discrepancies to truncation at kernel boundaries rather than to the FP32 interior, and uses teacher forcing to separate local deviations from autoregressive cascade; we reuse both insights in an INT8 setting where an exact-integer control is available. Hawkeye~\cite{badash2026hawkeye} reproduces tensor-core arithmetic on CPU and audits rounding, subnormal handling, and accumulation order at instruction level, establishing that low-level semantics are testable rather than noise. Kernel-contract work~\cite{xu2026contracts} provides vocabulary and divergence bounds between training and inference kernels; DiFR~\cite{karvonen2025difr} verifies deployments from fingerprints without mechanism isolation. We instantiate one empirically: shared-operand equality of the exact accumulator, an epilogue tolerance, and an intervention that must restore equality, each checkable on a concrete kernel pair.

\subsection{Quantized GEMM kernels}
Quantized-kernel research has mostly asked which implementation is \emph{fastest}, though not exclusively: Ootomo \emph{et al.}~\cite{ootomo2024dgemm} and Abdelfattah \emph{et al.}~\cite{abdelfattah2025intmm} ask instead what integer exactness buys numerically, emulating FP64 GEMM on integer matrix units, and derive the admissible reduction depth as a function of accumulator width. Their constraint is the same no-overflow condition in the log domain rather than as a magnitude bound; the instantiated thresholds differ between the two works and we claim shared structure with (\ref{eq:bound}), not numerical identity. LiquidGEMM~\cite{liquidgemm2025} engineers W4A8 GEMM for serving efficiency. Han and Qu~\cite{han2026matched} use a matched FP16 intermediate to separate a runtime ratio from a combined kernel-and-quantization ratio, treating that split as descriptive rather than causally independent. We ask a different question, whether two such implementations \emph{agree} numerically, holding the captured int8 operands, scales, shapes, and output dtype fixed so that only the scaled-MM implementation changes. Our earlier audit~\cite{chen2026specsheets} documents why the kernel selected for a format is itself a stack property; this paper measures what the selection changes.

\subsection{Concurrent work inside our review window}
Several 2026 preprints reach adjacent conclusions and appeared inside the review window declared in Section~\ref{sec:intro}; we position against them rather than claim priority over them. Where one of them makes an argument we also make, we say so and do not claim it as ours. Closest on the arithmetic is Cruz Romero and Maldonado Guerra~\cite{cruzromero2026arm}, who report that INT8 quantization collapses ARM dispatch variants to one equivalence class with byte-identical INT32 accumulators across dispatch paths --- the same localization argument on a different instruction set, without the epilogue decomposition or the pre-registered per-layer predictions. Closest on the transfer is MarginGate~\cite{chu2026margingate}, which triggers verification on low top-1 margin for batch-invariant serving; it uses the margin as an operational trigger, whereas we estimate and calibrate the margin-to-flip relation itself. Wu \emph{et al.}~\cite{wu2026decisions} predict which decisions low-bit quantization breaks, but sweep bit-width so their perturbation scales with it; our perturbation is a fixed sub-ulp epilogue difference at constant precision, so the two are not commensurable. On verification and identification, Cankaya~\cite{cankaya2026bitexact} verifies bit-exact inference, Wimbauer \emph{et al.}~\cite{wimbauer2026fingerprinting} solve the inverse problem of identifying which system produced an output, Gond \emph{et al.}~\cite{gond2026llm42} enforce determinism through verified speculation, and Fu \emph{et al.}~\cite{fu2026probabilities} expose nondeterminism at the token-probability level. Veit~\cite{veit2026speclang} proposes a specification language for kernel correctness across silicon; our conformance table is narrower and instantiated for one kernel pair.

\section{The Exactness Structure of a Scaled INT8 GEMM}
\label{sec:theory}

This section establishes what an implementation of a scaled INT8 GEMM is and is not free to do. We decompose the layer into three arithmetic stages, show that the first two admit no implementation freedom under stated conditions, and derive from the third an intervention that removes the freedom in the remainder. Each condition is testable, and Section~\ref{sec:boundary} tests it on the measurement hardware.

A W8A8 linear layer computes $Y = (A W^{\top}) \cdot s_a s_w$ where $A$ is an $M{\times}K$ int8 activation tile with per-token scales $s_a$, $W$ is an $N{\times}K$ int8 weight matrix with per-channel scales $s_w$, the product accumulates in INT32, and the scaled result rounds to bfloat16. Three elementary facts partition this pipeline into stages that \emph{can} and \emph{cannot} legally differ between implementations.

\textbf{(i) Exact, order-free accumulation.} With symmetric int8 quantization both operands lie in $[-127,127]$, so each product lies in $[-16129,16129]$, the bound our emulation uses; the unrestricted int8$\times$int8 extreme is $(-128)^2 = 16384$. For these checkpoints the operative bound is $16256$, and exactly so: the quantizer divides by $255/2 = 127.5$, which carries the \emph{weights} to $-128$, while the captured activations stay within $[-127,127]$ at all 196 layers. The largest product is thus $128 \times 127$; $16129$ does not apply because the weights do reach $-128$, and $16384$ is unreachable because the activations never do. Version~1 placed the $-128$ on the activation side and called $16256$ conservative: the number and everything derived from it stand, but its justification is a measurement, not a margin. With $16256$ the worst-case accumulator magnitude after $K$ terms obeys
\begin{equation}
\max_{i,j} |{\rm acc}_{ij}| \;\le\; 16256\,K \;<\; 2^{31}
\quad\text{for all } K \le 132{,}104 ,
\label{eq:bound}
\end{equation}
and $K \le 131{,}071$ even under the unrestricted $16384$, and the limit in (\ref{eq:bound}) is exact for the same reason. The layer depths here ($K \le 32{,}768$) satisfy (\ref{eq:bound}) with a factor of about four of margin on the worst case, about ten at the deepest layer actually present ($K = 12{,}288$); the accumulators actually observed are far smaller, the largest being $2{,}374{,}517$, some $900\times$ below $2^{31}$ and $2.82$ bits below the $2^{24}$ threshold discussed next. Integer addition is associative and commutative, and the integer tensor-core instructions accumulate into INT32 without intermediate rounding~\cite{ptxisa}. This is a property of the instruction path, not of int8 arithmetic in general: oneDNN documents that on AVX2 and AVX-512 without DL Boost its int8 GEMM accumulates pairs of products through a \emph{saturating} 16-bit intermediate, so the alibi does not extend to such paths without checking them~\cite{onednnint8}. On the instructions used here: given (\ref{eq:bound}), any tiling, any split-$K$, any reduction tree produces the identical INT32 value. The bound is not new as an analytic object: the FP64-emulation literature derives the admissible slice width and depth from the accumulator's bit budget~\cite{ootomo2024dgemm,abdelfattah2025intmm}, and accumulator-aware quantization makes overflow avoidance a design constraint with guarantees~\cite{colbert2023a2q,colbert2025axe}. What we add is its use as a \emph{control}: a per-layer, pre-registered prediction of where two kernels must agree bitwise. Order-independence follows from integer associativity, which the same sources already document. We verify this on hardware, including all-extreme worst-case matrices with $-128$ activations, in Section~\ref{sec:boundary}.

\textbf{(ii) The $2^{24}$ representability threshold.} Every integer accumulator is exactly representable in float32 when $|{\rm acc}| \le 2^{24}$. Above that threshold exact representability is no longer guaranteed, though individual values may still be exact: $2^{24}{+}1$ rounds whereas $2^{24}{+}2$ does not. A layer whose accumulators stay below the threshold on the evaluated inputs therefore enters the epilogue with no information loss, and this is checkable \emph{in advance} by an exact emulation (fp64 holds these integers exactly). This yields per-layer, falsifiable predictions of where bitwise agreement is required, before any kernel comparison runs.

The epilogue is where implementations do differ by design: fusing it is an explicit optimization target with its own compiler frameworks~\cite{chen2024evt}, so two kernels can fuse the same arithmetic differently while both remain correct in the conventional sense.

\textbf{(iii) Power-of-two commutation.} Multiplying a float32 value by $2^{k}$ shifts the exponent and leaves the mantissa untouched, so for finite normal results it commutes with rounding,
\begin{equation}
\mathrm{rnd}\!\left(x\right) \cdot 2^{k} \;=\; \mathrm{rnd}\!\left(x \cdot 2^{k}\right),
\label{eq:commute}
\end{equation}
whenever both sides are finite and normal. Constrained scale forms are themselves established: HAWQ-V3 uses dyadic scales of the form $b/2^{c}$ to keep the whole pipeline integer-only~\cite{yao2021hawqv3}, which is a weaker restriction than a power of two, and RAPQ fits power-of-two scales for accuracy at low bit-width~\cite{yao2022rapq}. In both the motivation is integer-only arithmetic or accuracy rather than cross-kernel agreement; we use the power-of-two restriction as a determinism intervention. If the weight scales are powers of two, (\ref{eq:commute}) makes the two epilogue orderings $(\mathrm{acc} \cdot s_a) \cdot s_w$ and $\mathrm{acc} \cdot (s_a s_w)$ produce bit-identical float32, and the final cast to bfloat16 then rounds the same value once in either arm. Equation (\ref{eq:commute}) requires finite normal values on both sides, no overflow or underflow, and matching rounding semantics; our unit tests exhibit its failure in the subnormal range (Section~\ref{sec:boundary}), so we state it as a conditional lemma, not a universal equivalence.

These three facts compose. Under (\ref{eq:bound}) and the $2^{24}$ condition the accumulator is above suspicion, which is the integer alibi. Real-scale differences can then arise only in the two scale multiplications and the output cast, each contributing at most rounding-level error, which places cross-kernel differences near one bfloat16 spacing. And by (\ref{eq:commute}) a power-of-two weight-scale probe removes even that freedom, making bitwise equality \emph{required} rather than merely expected. A floating-point GEMM has no analogous structure. Products of the e4m3 format~\cite{micikevicius2022fp8} are exact in float32, but their summation rounds at every step and is order-dependent, so two implementations may legitimately differ in ways that grow with reduction depth. This asymmetry is testable, and Section~\ref{sec:ksweep} tests it.

\section{Methods and Governance}
\label{sec:methods}

Because the central claim is that two implementations differ while everything else is held equal, the credibility of the study rests on how ``everything else'' is enforced and audited. This section describes the pinned stack, the pre-registration that fixed the estimators and the 1.7B predictions before measurement (the 8B list is pinned but not blind, and is treated separately), the manifest-level contract that verifies the isolation from the artifacts rather than from intent, and the estimator semantics. Pre-registration is not yet established practice in predictive modeling; it has been argued for as a remedy there~\cite{hofman2023prereg}, and the stronger discipline it approximates is blind analysis as practiced in experimental physics~\cite{klein2005blind}. We state below exactly where our protocol is pinned but not blind.

\textbf{Stack.} All measurements ran on one RTX 4090 (sm\_89, driver 580.173.02, clocks under default management) with vLLM~\cite{vllm} 0.27.1 pinned by container digest; the SGLang~\cite{sglang} arm is pinned by digest likewise. The arms ran back to back on an otherwise idle GPU under default clock management; clock state is not part of a bitwise comparison, but this replaces a pre-registered control and is listed among the deviations below. The model family is Qwen3~\cite{qwen3} at 1.7B and 8B, quantized to W8A8 INT8 (channel-wise symmetric weights, dynamic per-token activations) with llm-compressor 0.13.0~\cite{llmcompressor}; parent revisions and checkpoint digests are recorded in tracked manifests. The two treatment arms select vLLM's \code{CutlassInt8ScaledMMLinearKernel} or \code{TritonInt8ScaledMMLinearKernel}~\cite{cutlass,tillet2019triton} via \code{VLLM\_DISABLED\_KERNELS}; both arms share the engine's activation-quantization op: read from the pinned container, the two kernel classes issue the identical call \code{ops.scaled\_int8\_quant(x.contiguous(), i\_s, i\_zp, symmetric=symmetric)}, and the Triton class is a subclass of the CUTLASS one. This is source-level evidence; we did not additionally verify at run time that the two arms produce byte-identical quantized activations and per-token scales, so end-to-end differences remain formally attributable to the bundled kernel path rather than to the epilogue alone. The controlled layer experiments of Section~\ref{sec:layers}, which feed both kernels the same captured operands, are where epilogue localization is established. Prompts are 64 deterministic 320-token windows of WikiText-2 (pinned dataset revision); generation is greedy.

\textbf{Pre-registration.} The protocol was locked before any backend-comparison measurement: estimators, the 1.7B per-layer prediction list, acceptance criteria for synthetic validation, and the analysis plan, all pinned by SHA-256. The 8B prediction list came later and is pinned but not blind, as detailed under \emph{Replication timing} below. Every later change is an append-only amendment; post-measurement amendments are labeled as such, and analysis-freedom choices introduced along the way (near-zero thresholds, bin supports, bootstrap counts) are enumerated in the amendments rather than silently absorbed. Measurement semantics were themselves revised twice under adversarial review by an independent automated reviewer, which is not peer review but did produce concrete counterexamples: bitwise divergence was separated from finite-only ulp distance, and all fitted models were placed under a common objective. In every case the raw artifacts were preserved and re-analyzed under new file names, the counterexamples became regression tests, and no headline number moved beyond reporting precision.

\emph{Deviations.} Four pre-registered items are not reported here, and we list them rather than let the pre-registration imply otherwise. Locked GPU clocks were replaced by default clock management, as noted above. The pre-registered batch sweep $\{1,4,16\}$ and its integer-path batch-invariance prediction were not executed; all per-layer captures instead use a single prefill-like tile ($M{=}512$), so batch dependence is untested at layer level. The forward-and-reverse ordering rounds were not run. The task-level McNemar test had no data to apply to, because no task-level evaluation was performed: this paper measures agreement, not task quality.

\emph{Replication timing.} The 8B results are a pinned but not independent replication. Its prediction list and acceptance criterion were hash-pinned before the authoritative rerun reported here, yet an earlier run on the same 8B data had already been seen at that point. We therefore treat 1.7B as the pre-registered case and 8B as a same-data replication under a pinned criterion, and do not present the two as equivalent evidence.

\textbf{Identity and treatment contract.} Each teacher-forcing artifact carries a manifest with the full prompt-list SHA-256, rails SHA-256, checkpoint digest, quantization-config and tokenizer digests, parent revision, engine version, container digest, and, as treatment evidence, the arm label, the kernel class captured from the engine's own selection log at run time, the selection environment, and the log digest. The analysis-side join verifies each field separately and fails closed on any mismatch, malformed digest, or missing kernel evidence; analyses of unverified data require an explicit flag that is recorded in the output schema. For the final teacher-forced comparison the join passes every check, so \emph{for that comparison} ``the only declared difference is the kernel'' is an audited property of the artifacts rather than a statement of intent. The end-to-end token artifacts predate the contract and carry no manifest-level kernel-selection evidence; we report them as unaudited on that dimension.

\textbf{Estimators.} Bitwise divergence is bit-pattern inequality (signed zeros count); numerical ulp distance is computed on finite pairs only, as ordered bf16 bit-pattern distance; normalized error statistics use fp64 and finite-reference RMS with near-zero elements sub-reported separately. Token-level statistics treat the prompt as the clustering unit (cluster bootstrap, leave-one-prompt-out influence); average precision uses grouped ties; calibration uses merged quantile bins with a minimum support and Beta(1,1) smoothing, with decreasing-isotonic~\cite{barlow1972isotonic} regression as a pre-specified sensitivity analysis and Brier scores~\cite{brier1950} against a constant-prevalence baseline. Regression tests (72 checks) cover the counterexamples raised in review, and run in a digest-pinned clean container.

\section{Layer-Level Results: the Alibi at Work}
\label{sec:layers}

This section applies the alibi: per-layer predictions of where bitwise agreement is required, then identical operands through both kernels under two scale regimes. The power-of-two regime here is not the probe checkpoint of Section~\ref{sec:intervention}, and the difference bounds what this evidence covers: at layer level \emph{every} scale on both sides becomes a single constant, $2^{-9}$ for activations and $2^{-8}$ for weights, so (\ref{eq:commute}) is tested at one point in scale space repeated across layers rather than over the exponent range a checkpoint spans. The probe checkpoint instead maps each weight scale to its own nearest power of two, leaving activation quantization untouched. Version~1 named both ``the pow2 probe''.

\begin{table}[t]
\caption{Per-layer verification. ``Safe'' means every observed accumulator magnitude stayed below $2^{24}$ on the pinned prompts, so bitwise identity is required under power-of-two scales, and observed. The 1.7B prediction list was pre-registered before any kernel comparison; the 8B list was hash-pinned before the authoritative rerun but after an earlier result on the same data was known (Section~\ref{sec:methods}). Under real scales, every observed difference stays within one bf16 spacing.}
\label{tab:layers}
\centering
\begin{tabular}{lcc}
\toprule
 & Qwen3-1.7B & Qwen3-8B \\
\midrule
int8 linear layers & 196 & 252 \\
predicted bitwise-safe & 196 & 252 \\
min headroom to $2^{24}$ (bits) & 2.82 & 3.36 \\
capture tile $M$ (tokens/layer) & 512 & 512 \\
uniform pow2 scales: bit-identical layers & 196/196 & 252/252 \\
real scales: bit-identical layers & 8/196 & 10/252 \\
real scales: max finite ulp distance & 1 & 1 \\
elements beyond 1 ulp (model-wide) & 0 & 0 \\
non-finite outputs observed & 0 & 0 \\
\bottomrule
\end{tabular}
\end{table}

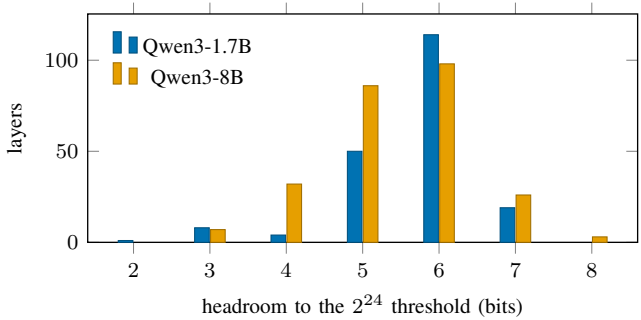
\begin{figure}[t]
\centering
\begin{tikzpicture}
\begin{axis}[
  width=\columnwidth, height=4.6cm,
  ybar, bar width=5.5pt,
  xlabel={headroom to the $2^{24}$ threshold (bits)},
  ylabel={layers},
  xmin=1.4, xmax=8.6,
  ymin=0,
  xtick={2,3,4,5,6,7,8},
  legend style={draw=none, fill=none, at={(0.03,0.95)}, anchor=north west, font=\footnotesize},
  tick label style={font=\footnotesize},
  label style={font=\footnotesize},
  axis line style={line width=0.5pt},
]
\addplot+[ybar, fill=okblue, draw=okblue!70!black, bar shift=-3pt] coordinates {(2,1) (3,8) (4,4) (5,50) (6,114) (7,19)};
\addplot+[ybar, fill=okorange, draw=okorange!70!black, bar shift=3pt] coordinates {(3,7) (4,32) (5,86) (6,98) (7,26) (8,3)};
\legend{Qwen3-1.7B, Qwen3-8B}
\end{axis}
\end{tikzpicture}
\caption{Observed accumulator headroom per layer (distance of the maximum $|{\rm acc}|$ below the $2^{24}$ float32 representability threshold, in bits; larger is safer). Every layer of both models clears the threshold, so the prediction lists mark all 448 layers bitwise-safe: their exact accumulators enter the epilogue without information loss. The 1.7B list was pre-registered before any kernel comparison; the 8B list was pinned before the authoritative rerun but is not blind (Section~\ref{sec:methods}). The claim is conditional on the pinned evaluation prompts; worst-case bounds do not clear the threshold.}
\label{fig:headroom}
\end{figure}

The prediction stage ran an exact fp64 emulation of the integer pipeline over the pinned prompts and recorded, per layer, the maximum accumulator magnitude. All 196 layers of the 1.7B model and all 252 of the 8B stayed below $2^{24}$, with at least 2.8 bits of headroom (Figure~\ref{fig:headroom}). The 1.7B list was hash-pinned before any kernel comparison ran; the 8B list was hash-pinned before the authoritative rerun reported here, at which point an earlier result on the same 8B data was already known. The verification stage then fed \emph{identical} captured int8 operands to both kernels, twice per layer: once with scales overridden to powers of two, once with the checkpoint's real scales. All captures use a single prefill-like tile of $M{=}512$ tokens per layer; the decode regime ($M{=}1$) is not covered at layer level.

Table~\ref{tab:layers} summarizes the outcome. Under the pow2 probe, where (\ref{eq:commute}) predicts bitwise identity for finite normal outputs, all 448 layers are bit-identical across CUTLASS and Triton: the lemma's predictions survived at every layer tested, and the alibi caught no epilogue bug in this kernel pair. We recorded no non-finite outputs (Table~\ref{tab:layers}) but did not separately count subnormal outputs, the one regime where the lemma is known to fail (Section~\ref{sec:boundary}). We also report no fault-injection or mutation test, so the sensitivity of each check to a genuinely faulty kernel is unmeasured: these are checks that passed, not checks demonstrated to fire. Under real scales, differences appear in 188 of 196 layers of the 1.7B model (the eight identical layers are small-$N$ projections where rounding coincidence survives), yet the largest difference anywhere in either model is a single bfloat16 spacing, and no element exceeds one ulp. The divergence is thus pervasive, rare per element (fractions around $5\times10^{-6}$), tightly bounded, and, by the alibi, attributable to the epilogue's scale multiplications and output rounding rather than to accumulation.

At the isolated-operator level the same pattern holds with random operands: identical inputs produce bit-identical outputs under pow2 scales at every shape tested, while real scales yield a handful of one-ulp differences out of half a million elements at $M{=}256$. The smaller shapes, including $M{=}1$, showed no differences, but at $2\times10^{3}$ output elements such a probe has essentially no power against a per-element rate near $5\times10^{-6}$, so it is not evidence of decode-regime agreement. A separate 200{,}000-sample experiment on the epilogue arithmetic alone, comparing the two multiplication orderings directly, produced zero bf16-visible differences (95\% binomial upper bound $1.5\times10^{-5}$), consistent with the per-layer rates.

\section{Boundary Validation}
\label{sec:boundary}

The claims above lean on three arithmetic facts, so we test the facts on the measurement hardware, in the pinned container, with an environment-stamped record. The $2^{24}$ boundary behaves exactly as IEEE 754~\cite{ieee754} prescribes: $\pm(2^{24}{-}1)$ and $\pm2^{24}$ round-trip int$\to$float32 exactly, $\pm(2^{24}{+}1)$ does not, $\pm(2^{24}{+}2)$ does. Integer accumulation is exact against an INT64 reference for random and for all-extreme matrices, including activations at $-128$ where the product bound is $16256$, at $K$ up to 32{,}768, on the GPU integer path (\code{torch.\_int\_mm}) and the fp64 emulation alike. Power-of-two commutation holds bitwise for finite normal values and both signed zeros across all tested exponents, and \emph{fails} in the subnormal range, exactly as the conditional lemma anticipates; true bfloat16 rounding midpoints, probed at one float32 ulp on either side, commute in every case. These tests bound the theory's jurisdiction: within it, the alibi is safe to use; outside it (subnormals, overflow), we make no claim.

\section{Divergence Structure versus Reduction Depth}
\label{sec:ksweep}

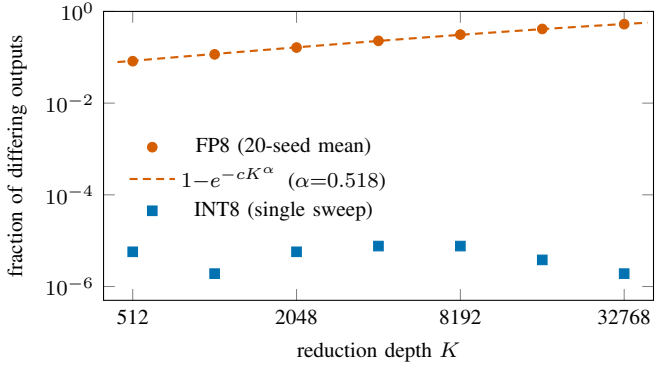
\begin{figure}[t]
\centering
\begin{tikzpicture}
\begin{loglogaxis}[
  width=\columnwidth, height=5.4cm,
  xlabel={reduction depth $K$},
  ylabel={fraction of differing outputs},
  xmin=400, xmax=42000,
  ymin=5e-7, ymax=1,
  xtick={512,2048,8192,32768},
  xticklabels={512,2048,8192,32768},
  legend style={draw=none, fill=none, at={(0.03,0.42)}, anchor=west, font=\footnotesize},
  tick label style={font=\footnotesize},
  label style={font=\footnotesize},
]
\addplot[only marks, mark=*, mark size=1.8pt, color=okverm] coordinates {
(512,0.0821) (1024,0.11569) (2048,0.16297) (4096,0.22818) (8192,0.31218) (16384,0.4142) (32768,0.52729)};
\addplot[no marks, densely dashed, thick, color=okverm, domain=450:40000, samples=120] {1-exp(-0.00345734*x^0.5184)};
\addplot[only marks, mark=square*, mark size=1.8pt, color=okblue] coordinates {
(512,5.72e-06) (1024,1.91e-06) (2048,5.72e-06) (4096,7.63e-06) (8192,7.63e-06) (16384,3.81e-06) (32768,1.91e-06)};
\legend{FP8 (20-seed mean), $1{-}e^{-cK^{\alpha}}$\, ($\alpha{=}0.518$), INT8 (single sweep)}
\end{loglogaxis}
\end{tikzpicture}
\caption{Cross-implementation divergence versus reduction depth $K$; both series use $M{=}256$, $N{=}2048$ and the same $K$ grid, but their designs otherwise differ and are not matched. FP8 (CUTLASS vs.\ \code{torch.\_scaled\_mm}): 20 seeds, nested-prefix inputs so that only $K$ varies within a seed, fixed per-tensor scalar scales, Gaussian operands cast to fp8; the fraction grows from 8\% to 53\% of elements and is best fit, among four pre-listed candidates under a common objective with held-out folds, by generalized saturation with exponent $\alpha=0.518$ (seed-bootstrap CI90 $[0.5181,0.5187]$; jackknife and leave-one-$K$-out ranges $[0.516,0.529]$), a shape \emph{compatible with} square-root-driven boundary crossing under saturation. INT8 (CUTLASS vs.\ Triton): a single unreplicated sweep in which uniform int8 operands and random per-token and per-channel scales are redrawn at each $K$; 18 differing elements in total ($1.9$--$7.6$ ppm), showing no growth detectable at this sample size. The exact accumulator leaves only per-element epilogue rounding, which does not accumulate with depth; the INT8 series tests that expectation weakly, and the per-layer results of Section~\ref{sec:layers} carry the stronger evidence.}
\label{fig:ksweep}
\end{figure}

If the alibi's account is right, the INT8 path cannot accumulate reduction-order rounding with depth: each element carries a fixed number of epilogue roundings regardless of $K$. Prevalence could still vary with $K$ if the distribution of accumulator and scaled values relative to the bf16 grid changes, so the prediction is the absence of an accumulation mechanism, not a theorem that prevalence must be constant. A float-accumulator format, by contrast, should show depth-dependent divergence, its order-dependent rounding crossing the output grid more often as $K$ grows. Figure~\ref{fig:ksweep} shows both signatures on the same axes. Across $K = 512$ to $32{,}768$ the INT8 arm produced 18 differing elements in total, one to four per grid point out of $524{,}288$ ($1.9$ to $7.6$ parts per million), in a single unreplicated sweep. A constant-rate model is entirely consistent with these counts ($\chi^2 = 3.8$, six degrees of freedom), but 18 events cannot exclude a trend of a factor of two either; we therefore report the INT8 series as showing no detectable growth at this sample size rather than as demonstrably flat. The FP8 fraction rises from 8.2\% to 52.7\%; among four pre-listed models fitted under a single objective (with $p{=}0$ rows retained and $\alpha$ free, never preset), generalized saturation $1-\exp(-cK^{\alpha})$ attains the lowest training and held-out error with $\alpha \approx 0.518$, stable under seed jackknife, leave-one-$K$-out, and disjoint half-range refits. We state this as compatibility with a square-root-shaped mechanism, not as a law: seven grid points on one implementation pair do not exclude other functional families.

The two formats also differ in how large the differences are, not only how often they occur. The median relative difference among differing elements sits near one bf16 spacing in both formats ($5.4\times10^{-3}$ to $6.5\times10^{-3}$ for FP8 across the sweep). Only the INT8 arm, however, keeps its whole distribution there: its maximum relative difference never exceeds one spacing at any $K$ ($\le 7.4\times10^{-3}$), and per-layer verification measures a maximum bf16 ulp distance of exactly one. In the FP8 arm the tail grows with depth as well. Among differing non-near-zero elements, the share at ulp distance $\ge 2$ rises from $8\%$ at $K{=}512$ to $33\%$ at $K{=}32{,}768$ ($0.6\%$ to $16.9\%$ of all output elements), the largest relative difference among them grows from $9\%$ to $125\%$ (seed means), and the maximum ulp distance is four orders of magnitude above one throughout. For FP8, then, both prevalence and magnitude grow with $K$; for INT8 no growth was detected, and the observed magnitude remained within one bf16 spacing throughout.

\section{From Ulps to Tokens}
\label{sec:tokens}

A one-ulp layer difference matters only if it changes a decision; this section measures that transfer at the sequence level, at single steps under teacher forcing, and as a function of the logit margin.

\begin{table}[t]
\caption{End-to-end greedy agreement; ``identical'' means the full token sequence matches bitwise. Continuations are 64 tokens except the 64-prompt rail run, which generates 256. At 1.7B each arm is self-consistent across two cold restarts; the 8B and 64-prompt runs were not repeated. Arms disagree with each other under real scales and agree bitwise under the pow2 probe checkpoint.}
\label{tab:e2e}
\centering
\begin{tabular}{lcc}
\toprule
comparison & 1.7B & 8B \\
\midrule
same arm, two cold runs (each arm) & 8/8 & --- \\
CUTLASS vs.\ Triton, real scales & 0/8 & 0/16 \\
\quad same, 64-prompt rail run & 0/64 & --- \\
CUTLASS vs.\ Triton, pow2 probe & 8/8 & 16/16 \\
SGLang vs.\ vLLM arms, real scales & 0/8 & --- \\
SGLang vs.\ vLLM, pow2 probe & 1/8 & --- \\
\bottomrule
\end{tabular}
\end{table}

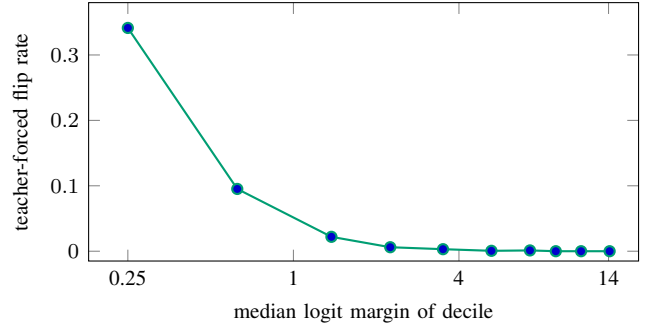
\begin{figure}[t]
\centering
\begin{tikzpicture}
\begin{semilogxaxis}[
  width=\columnwidth, height=5.0cm,
  xlabel={median logit margin of decile},
  ylabel={teacher-forced flip rate},
  xmin=0.18, xmax=18,
  ymin=-0.015, ymax=0.38,
  xtick={0.25,1,4,14},
  xticklabels={0.25,1,4,14},
  tick label style={font=\footnotesize},
  label style={font=\footnotesize},
]
\addplot+[mark=*, mark size=1.9pt, thick, color=okgreen] coordinates {
(0.25,0.3413) (0.625,0.0952) (1.375,0.022) (2.25,0.0061) (3.5,0.0031)
(5.25,0.0006) (7.25,0.0012) (9.0,0.0) (11.125,0.0) (14.125,0.0)};
\end{semilogxaxis}
\end{tikzpicture}
\caption{Cross-kernel token flips against the logit margin, teacher-forced on identical contexts (64 prompts $\times$ 256 positions, identity-verified artifacts). Flip rate falls monotonically from 34\% in the lowest-margin decile to zero above a margin of ${\sim}9$; overall 769 of 16{,}384 positions flip (4.7\%). The margin predicts flips with ROC-AUC 0.942 (prompt-cluster CI90 $[0.935, 0.949]$); a margin-binned calibration achieves held-out Brier 0.0352 against a 0.0440 constant-prevalence baseline.}
\label{fig:margin}
\end{figure}

Table~\ref{tab:e2e} gives the sequence-level picture. Under real scales the two kernel arms agree on no sequence at either model size, while each arm reproduces itself exactly across cold restarts; adding SGLang as a third implementation yields three mutually distinct deterministic functions. Teacher-forced replay on identical contexts (which removes autoregressive cascade) shows how one-ulp layer differences become token differences: single-position flips occur at 4.7\% of positions and concentrate almost entirely at small logit margins (Figure~\ref{fig:margin}). The margin alone ranks flip risk with ROC-AUC 0.942, and a calibrated margin-to-flip-probability map beats a constant baseline on held-out prompts. Once any position flips, contexts diverge and full sequences separate, which is why sequence-level agreement is binary in practice.

Two honest negatives accompany the transfer analysis. First, position-exact prediction fails: the first teacher-forced flip position does not match the first free-running divergence position, because replay scores all positions in one prefill pass while generation ran incrementally, and the noise draws differ between those regimes. The transfer claim holds at the distribution level, not per position. Second, the same-arm prefill-versus-decode comparison flips 8.6\% of rail tokens under real scales, and the pow2 probe reduces but does not eliminate it (2.9\% residual). That residual originates outside the controlled INT8 scaled-MM epilogue; our first-divergence instrumentation did not reach module attribution (its engine glue fails before capture, and we did not guess), so we report the 2.9\% as \emph{unresolved} rather than attributing it to attention, KV-cache handling, normalization, or the unquantized head.

\section{The Pow2 Intervention End to End}
\label{sec:intervention}

The power-of-two probe converts the localization into a counterfactual. Rewriting all 573{,}440 weight-scale values of the 1.7B checkpoint (and correspondingly for 8B) to the nearest power of two, leaving weights, activation quantization, prompts, and both kernel arms untouched, restores \emph{bitwise} end-to-end agreement: 8/8 sequences at 1.7B and 16/16 at 8B (Table~\ref{tab:e2e}). Removing the suspected freedom removes the divergence, which is the strongest evidence this study offers that real-scale cross-kernel differences originate in floating-point scale application rather than anywhere upstream.

Three scope notes temper the result. The probe is a conformance instrument, not a deployment recipe: rounding scales to powers of two perturbs the model, and we measured no accuracy, calibration, or throughput consequences of doing so in production. The intervention repairs \emph{kernel} interchangeability, not \emph{engine} interchangeability: SGLang still disagrees with the vLLM arms under the probe (engines differ in attention, sampling, and more, beyond the linear kernels). And the end-to-end probe runs used one engine-process configuration, recorded in the manifests, to make kernel-selection evidence capturable; the raw-scale baselines were generated under the engine's default process mode.

\section{Discussion}
\label{sec:discussion}

\textbf{A conformance procedure.} The controls in this paper compose into an artifact-backed procedure, instantiated by the scripts of this study for one kernel pair and adaptable to another; Table~\ref{tab:conformance} states it as seven checks, each with what a violation points to first and the conditions under which its underlying argument does not apply. Two properties make the procedure usable rather than aspirational. The first five checks have falsifiable outcomes rather than thresholds to tune, the accumulator check admitting no legal difference at all and the power-of-two check turning a rounding argument into a required equality; the last two are explicitly tolerance-based and must be set per workload. And the identity contract makes ``same operands, declared treatment only'' an audited property of the artifacts, so a passing run cannot be an accident of provenance. Our regression tests cover the estimator and contract counterexamples that adversarial review produced; the GPU scripts instantiate the arithmetic probes for this kernel pair rather than providing a general runner.

\begin{table*}[t]
\caption{The conformance procedure implied by the controls in this paper. The scripts of this study instantiate these checks for one scaled-INT8 kernel pair; another pair would require adapting them rather than running them unchanged. Each row names what a violation points to first; the exclusions are the conditions under which the underlying argument does not apply.}
\label{tab:conformance}
\centering
\footnotesize
\begin{tabular}{@{}>{\raggedright\arraybackslash}p{3.0cm}>{\raggedright\arraybackslash}p{5.6cm}>{\raggedright\arraybackslash}p{7.4cm}@{}}
\toprule
Check & Requirement & A violation suggests, or first check \\
\midrule
Shared operands & Both arms consume the same int8 tensors and scales, verified from artifacts & The comparison is not kernel-only; no localization is licensed \\
INT32 no-overflow & $16256K < 2^{31}$ for the layer's $K$ & Accumulation may wrap, so the integer alibi does not apply at all and no localization is licensed \\
Lossless FP32 entry & Observed $|{\rm acc}| \le 2^{24}$ per layer & Only the universal no-loss guarantee at the epilogue boundary is lost; individual values may still be exact, so judge per value or widen the tolerance rather than discard the layer \\
Exact accumulator & Bit-identical INT32 accumulation across arms & An integer-path defect: the arithmetic admits no legal difference \\
Pow2-scale identity & Bit-identical outputs under power-of-two scales, excluding subnormal and overflow results & First check the listed preconditions and each arm's intermediate semantics; if those hold, an epilogue defect, since rounding order cannot matter under (\ref{eq:commute}) \\
Real-scale tolerance & Cross-arm bf16 ulp distance $\le 1$, assuming both arms round once from fp32 intermediates; near-zero and non-finite elements reported separately & Suggests reduced intermediate precision, fusion, or another departure from single-rounding semantics rather than mere double rounding \\
Token-level risk & Flip risk \emph{ranked and calibrated} from the margin distribution of the intended workload (ROC-AUC 0.94, held-out Brier 0.035 here); no bound is implied & The margin does not carry flip risk on that workload; recalibrate before using it as a gate \\
\bottomrule
\end{tabular}
\end{table*}

\textbf{What the contrast teaches.} An exact, order-free accumulator confines implementation freedom to a fixed number of per-element epilogue roundings and cannot accumulate reduction-order rounding at all; a float accumulator lets that freedom grow with reduction depth, in prevalence and in magnitude alike. Format choice therefore determines which \emph{class} of reproducibility contract is available at all. For integer paths, near-bitwise interchangeability is at least specifiable and testable at rounding-level tolerance; for float paths, batch-invariant engineering~\cite{he2025batchinvariance} or reduction-tree alignment~\cite{tbik2025} pays throughput for a guarantee the algebra does not give for free.

\textbf{Reproducibility risk versus quality.} Token-level disagreement is a reproducibility and auditability problem (evaluation drift, cache invalidation, provider verification~\cite{karvonen2025difr}), not automatically a quality problem; texts that differ bitwise may be equally good. Our margin analysis quantifies risk of divergence, and we make no claim about task-quality degradation.

\section{Limitations}
\label{sec:limits}

(1)~\emph{Scope of the causal evidence.} One consumer GPU (sm\_89), one model family at two sizes, one engine version, and two INT8 kernel implementations (plus one FP8 pair and one cross-engine comparison). Nothing here licenses claims across GPU generations, model families, or future kernels. (2)~\emph{The 2.9\% residual is unresolved.} Same-arm regime variance under the probe is reported without module attribution; the first-divergence instrumentation reached its engine glue and stopped there. (3)~\emph{The probe is not a validated mitigation.} No accuracy, calibration, or performance evaluation of pow2-rounded scales was performed. (4)~\emph{Whole-model inference carries no bitwise guarantee.} The alibi covers the integer GEMM stage only; attention, normalization, rotary embeddings, and the unquantized head remain floating point and lie outside its jurisdiction. The prefill-versus-decode variance under the probe shows that at least one source outside the controlled INT8 epilogue is active; which module it is remains unattributed (limitation 2), and we do not claim it is any of those just named. (5)~\emph{Environment rebuild is partial.} Tests ran in a digest-pinned container with an exact-version closure, but without wheel hashes or a published image. The test added at the final re-pin was verified locally rather than in a fresh clean-container record, and the Dockerfile copies a \code{fixtures/} directory absent from the published tree, so the container cannot be rebuilt from the released checkout alone. The historical checkpoints also lack calibration-dataset revisions, a provenance gap at quantization time. (6)~\emph{No positive control.} None of the conformance checks has been exercised against a deliberately faulty kernel, so their sensitivity and false-negative rates are unmeasured; every check reported here passed. (7)~\emph{Regime scope.} Layer-level localization uses a single prefill-like tile ($M{=}512$); the decode regime is untested at layer level, and the isolated-operator probe at $M{=}1$ is too small to detect the observed per-element rate. (8)~\emph{Statistical scope.} FP8 depth-dependence is a compatibility statement from seven grid points on one implementation pair; token statistics condition on the pinned prompts and greedy decoding.

\section{Conclusion}

Two deterministic implementations of the same quantized linear layer need not be the same function, and in a pinned, audited configuration they are not: in 188 of 196 layers at least one output element differs, every differing finite element by at most one bfloat16 ulp, and no greedy sequence matches. The integer alibi turns that observation into a mechanism: an exact, order-free accumulator acquits the reduction stage, per-layer representability predictions say where bitwise agreement is mandatory, and a power-of-two scale intervention makes the remaining freedom vanish, layer by layer and end to end. The result is not that quantized inference is fragile, but that for integer paths interchangeability is \emph{checkable}: a conformance contract with an exact control, rounding-level tolerances, and a counterfactual repair, executable before deployment rather than discovered after it.

For practitioners the reading is narrow but actionable. Record which kernel served a result: the choice is a load-time decision that no model card reports and that our measurements show is outcome-relevant, so evaluation numbers are conditioned on it. Where bitwise reproducibility across implementations is a requirement rather than a preference, the integer path is where that requirement is at least \emph{specifiable and testable}, and Table~\ref{tab:conformance} is the check to run before substituting one implementation for another. It is not yet a guarantee: under real scales our arms differed in 188 of 196 layers and on every sequence compared, and bitwise agreement appeared only under the probe, which we have not evaluated as a deployment configuration (limitation 3). Where reproducibility is not a requirement, the margin analysis supplies the quantity to bound instead: not whether outputs differ, but how often a difference reaches a decision boundary.

Three openings remain. The residual same-arm variance under the probe needs first-divergence instrumentation before it can be attributed rather than reported. The power-of-two intervention needs an accuracy, calibration, and throughput evaluation before anyone proposes it as a mitigation. And the exact-versus-floating accumulator contrast deserves replication across GPU generations, model families, and the four-bit formats, where the epilogue carries even more of the arithmetic.

\section*{Artifact Statement}

The pre-registration (locked before backend-comparison measurements; nine append-only amendments), per-layer prediction lists, all measurement artifacts with SHA-256 digests, per-run manifests including kernel-selection evidence, the identity-contract implementation with its regression tests (72 checks; clean-container record), and the analysis code are maintained in a version-controlled repository with a linear commit history. The final validation iterations used separate plan-and-code pins and result commits; earlier exceptions, including one batch that published plan, code, and first results together and one execution commit that carried a semantic fix, are recorded append-only in the governance log rather than smoothed over. The repository is \url{https://github.com/luka-krixvon/integer-alibi}; it is private at the time of writing and will be made public with the announcement of this paper, at which point the manuscript and result commit hashes will be listed here. Known reproducibility gaps are enumerated in Section~\ref{sec:limits}.

\bibliographystyle{IEEEtran}
\bibliography{refs}

\end{document}